\documentclass[nohyperref]{article}

\usepackage{amsmath,amsfonts,bm}

\def\eqref#1{equation~\ref{#1}}

\def\1{\bm{1}}

\DeclareMathAlphabet{\mathsfit}{\encodingdefault}{\sfdefault}{m}{sl}
\SetMathAlphabet{\mathsfit}{bold}{\encodingdefault}{\sfdefault}{bx}{n}

\usepackage{microtype}
\usepackage{graphicx}
\usepackage{subfigure}
\usepackage{booktabs} 

\usepackage{hyperref}
\usepackage{pifont}

\usepackage[dvipsnames]{xcolor}
\definecolor{Tianlong_color}{rgb}{0.858, 0.188, 0.478}

\usepackage[accepted]{icml2022}

\usepackage{amsmath}
\usepackage{amssymb}
\usepackage{mathtools}
\usepackage{amsthm}
\usepackage{makecell}
\usepackage{diagbox}
\usepackage{subfigure}
\usepackage{multirow}
\usepackage{multicol}
\usepackage{varwidth}
\usepackage{adjustbox}
\usepackage{threeparttable}
\usepackage{enumitem}
\usepackage{pifont}
\usepackage{xspace}
\usepackage{algorithmic}
\renewcommand{\algorithmicrequire}{\textbf{Input:}}
\renewcommand{\algorithmicensure}{\textbf{Output:}}
\newcommand{\cmark}{\textcolor{green}{\ding{51}}\xspace}%

\usepackage[capitalize,noabbrev]{cleveref}

\theoremstyle{plain}

\theoremstyle{definition}

\theoremstyle{remark}

\usepackage[textsize=tiny]{todonotes}

\icmltitlerunning{Data-Efficient Double-Win Lottery Tickets from Robust Pre-training}

\begin{document}

\twocolumn[
\icmltitle{Data-Efficient Double-Win Lottery Tickets from Robust Pre-training}




\begin{icmlauthorlist}
\icmlauthor{Tianlong Chen}{ut}
\icmlauthor{Zhenyu Zhang}{ut}
\icmlauthor{Sijia Liu}{msu,ibm}
\icmlauthor{Yang Zhang}{ibm}
\icmlauthor{Shiyu Chang}{ucsb}
\icmlauthor{Zhangyang Wang}{ut}
\end{icmlauthorlist}

\icmlaffiliation{ut}{Department of Electrical and Computer Engineering, University of Texas at Austin}
\icmlaffiliation{msu}{Michigan State University}
\icmlaffiliation{ibm}{MIT-IBM Watson AI Lab}
\icmlaffiliation{ucsb}{University of California, Santa Barbara}

\icmlcorrespondingauthor{Zhangyang Wang}{atlaswang@utexas.edu}

\icmlkeywords{Machine Learning, ICML}

\vskip 0.3in
]



\printAffiliationsAndNotice{}  

\begin{abstract}
\vspace{-2mm}
Pre-training serves as a broadly adopted starting point for transfer learning on various downstream tasks. Recent investigations of \textit{lottery tickets hypothesis} (LTH) demonstrate such enormous pre-trained models can be replaced by extremely sparse subnetworks (a.k.a. \textit{matching subnetworks}) without sacrificing transferability. However, practical security-crucial applications usually pose more challenging requirements beyond standard transfer, which also demand these subnetworks to overcome adversarial vulnerability. In this paper, we formulate a more rigorous concept, \textbf{Double-Win Lottery Tickets}, in which a located subnetwork from a pre-trained model can be independently transferred on diverse downstream tasks, to reach \textbf{BOTH} the same standard and robust generalization, under \textbf{BOTH} standard and adversarial training regimes, as the full pre-trained model can do. We comprehensively examine various pre-training mechanisms and find that robust pre-training tends to craft sparser double-win lottery tickets with superior performance over the standard counterparts. For example, on downstream CIFAR-10/100 datasets, we identify double-win matching subnetworks with the standard, fast adversarial, and adversarial pre-training from ImageNet, at $89.26\%/73.79\%$, $89.26\%/79.03\%$, and $91.41\%/83.22\%$ sparsity, respectively. Furthermore, we observe the obtained double-win lottery tickets can be more \textbf{data-efficient} to transfer, under practical data-limited (e.g., $1\%$ and $10\%$) downstream schemes. Our results show that the benefits from robust pre-training are amplified by the lottery ticket scheme, as well as the data-limited transfer setting. Codes are available at {\small \url{https://github.com/VITA-Group/Double-Win-LTH}}. 
\end{abstract}

\section{Introduction}
\begin{figure}
    \centering
    \includegraphics[width=0.98\linewidth]{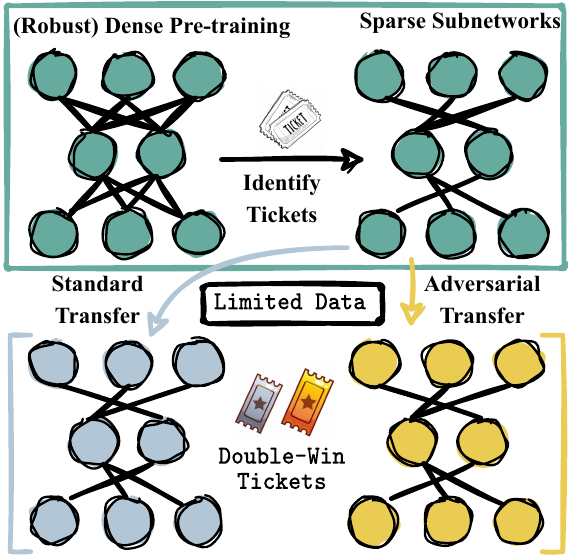}
    \vspace{-4mm}
    \caption{\small Overview of our work paradigm: we investigate the existence of double-win lottery tickets drawn from robust pre-training in the scenario of transfer learning, with the full training data and the limited training being available, respectively.}
    \vspace{-5mm}
    \label{fig:teaser}
\end{figure}

\begin{figure*}[!htb]
    \centering
    \vspace{-2mm}
    \includegraphics[width=1.0\linewidth]{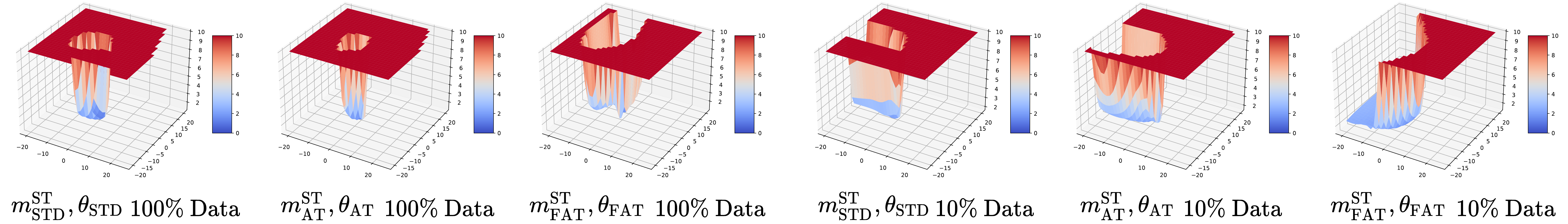}
    \vspace{-8mm}
    \caption{{\small Loss landscape visualization of subnetworks ($73.79\%$ sparsity) from diverse adversarial fine-tuning schemes. Each sparse network is first identified by IMP and standard training on the pre-training task with standard $\theta_{\mathrm{STD}}$, fast adversarial $\theta_{\mathrm{FAT}}$, or adversarial $\theta_{\mathrm{AT}}$ pre-training, respectively. Then, they are fine-tuned from the corresponding (robust) pre-training on downstream CIFAR-10 with $100\%$ or $10\%$ training data.
    }}
    \vspace{-5mm}
    \label{fig:3d_loss}
\end{figure*}

The lottery tickets hypothesis (LTH)~\cite{frankle2018lottery} demonstrates that there exist subnetworks in dense neural networks, which can be trained in isolation from the same random initialization and match the performance of the dense counterpart. We call such subnetworks as winning tickets. Unlike the conventional pipeline~\cite{han2015deep} of model compression that follows the train-compress-retrain process and aims for efficient inference, the LTH sheds light on the potential for more computational savings by training a small subnetwork from the start if only we had known which subnetwork to choose. However, finding these intriguing subnetworks is quite costly since the current most effective approach, iterative magnitude pruning (IMP)~\cite{frankle2018lottery,han2015deep}, requires multiple rounds of burdensome (re-)training, especially for large models like BERT~\cite{devlin2018bert}. Fortunately, recent studies~\cite{chen2020lottery,chen2020lottery2} provide a remedy by leveraging the popular paradigm of pre-training and fine-tuning, which first identifies critical subnetworks (a.k.a. pre-trained tickets) from standard pre-training and then transfers to a range of downstream tasks. The demonstrated \textit{universal transferability} across various datasets and tasks, shows the positive sign of replacing the gigantic pre-trained models with a much smaller subnetwork while maintaining the impressive downstream performance and leading to substantial memory/computation reductions. Meantime, the extraordinary cost of both pre-training and finding pre-trained tickets can be amortized by reusing and transferring to diverse downstream tasks.

Nevertheless, in practical settings, the deployed models usually ask for strong robustness, which is beyond the scope of standard transfer, e.g., for safety-critical applications like autonomous cars and face recognition. Therefore, a more challenging requirement arises, which demands the located subnetworks can effectively transfer in both standard and adversarial training schemes~\cite{madry2017towards}. Thus,
it is a new perspective to investigate the transferability of pre-trained tickets across diverse training regimes, differing from previous works~\cite{chen2020lottery,chen2020lottery2} on transferring downstream datasets and tasks. This inspires us to propose a new hypothesis of lottery tickets.
Specifically, when an identified sparse subnetwork from pre-training can be independently trained (transferred) on diverse downstream tasks, to match the same accuracy and robustness, under both standard and adversarial training regimes, as the full pre-trained model can do -- we name it a \textbf{Double-Win Lottery Ticket} illustrated in Figure.~\ref{fig:teaser}.


Meanwhile, inspired by~\cite{salman2020adversarially}, which suggests that robust pre-training shows better transferability for dense models, we examine (1) whether this appealing property still holds under the lens of sparsity; (2) how can robust pre-training benefits our double-win tickets compared to its standard counterpart. To address such curiosity, we comprehensively investigate representative robust pre-training approaches besides standard training, including fast adversarial (FAT)~\cite{wong2020fast} and adversarial (AT)~\cite{madry2017towards} pre-training. Our results reveal the prevailing existence of double-win tickets with different pre-training, and suggest the subnetworks obtain from AT pre-trained models consistently achieve superior generalization and robustness, under both standard and adversarial transfer learning, when the typical full training data is available for downstream tasks.

Yet, another critical constraint in real-world scenarios is the possible scarcity of training data (e.g., due to the difficulty of data collection and annotation).
What makes it worse is that satisfactory adversarial robustness intrinsically needs more training samples~\cite{schmidt2018adversarially}. Our proposed double-win tickets from robust pre-training tackle this issue by leveraging the crafted sparse patterns as an inductive prior, which ($i$) is found to reduce the sample complexity~\cite{zhang2021why} and brings data efficiency~\cite{chen2021ultra}; ($ii$) converges to a flatter loss landscapes with improved robust generalization as advocated by~\cite{wu2020revisiting,hein2017formal}, particularly for data-scarce settings shown in Figure~\ref{fig:3d_loss}. To support these intuitions, extensive experiments about few-shot (or data-efficient) transferability are evaluated with only $10$\% or $1$\% data for adversarial downstream training~\cite{jiang2020robust}. In what follows, we summarize our \textbf{contributions} in order to bridge LTH and its practical usage in the data-limited and security-crucial applications: 
\vspace{-2mm}


\begin{itemize}
    \item We define a more rigorous notion of double-win lottery tickets, which requires the sparse subnetworks found on pre-trained models to have the same transferability as the dense pre-trained ones: in terms of both accuracy and robustness, under both standard and adversarial training regimes, and towards a variety of downstream tasks. We show such tickets widely exist.
    \item Using IMP, we find double-win tickets broadly across diverse downstream datasets and at non-trivial sparsity levels $79.03\% \sim 89.26\%$ and $83.22\% \sim 96.48\%$ sparsity, using the fast adversarial (FAT) and adversarial (AT) pre-training. In general, subnetworks located from the AT pre-trained model have superior performance than FAT and standard pre-training.
    \item We further demonstrate the intriguing property of double-win tickets in the data-limited transfer settings (e.g., $10$\%, $1$\%). In this specific situation, FAT can surprisingly find higher-quality subnetworks with small sparsity while AT overtakes in a larger sparsity range.
    \item We show that adopting standard or adversarial training in the process of IMP makes no significant difference for the transferability of identified subnetworks on downstream tasks.
\end{itemize}

\section{Related Works} 

\paragraph{The lottery tickets hypothesis (LTH).} The (LTH)~\cite{frankle2018lottery} points out the existence of sparse subnetworks which are capable of training from scratch and match or even surpass the performance of the full network. \cite{frankle2019stabilizing, renda2020comparing} further scale up LTH to larger datasets and networks by weight rewinding techniques that re-initialize the subnetworks to the weight from the early training stage instead of scratch. Follow-up researchers have explored LTH in various fields, including image classification~\cite{frankle2018lottery,liu2019rethinking,Wang2020Picking,evci2019difficulty,ma2021good,You2020Drawing}, natural language processing~\cite{gale2019state,yu2019playing,chen2020earlybert,chen2020lottery}, vision+language multi-modal tasks~\cite{gan2021playing}, graph neural networks~\cite{chen2021unified}, generative adversarial networks~\cite{chen2021gans,chen2021ultra}, reinforcement learning~\cite{yu2019playing} and life-long learning~\cite{chen2021long}. Most existing works of LTH identify subnetworks by resource-consuming (iterative) weight magnitude pruning~\cite{han2015deep,frankle2018lottery}. Studies about the transferability of the subnetworks provide a potential offset to the computationally expensive process of finding high-quality subnetworks. \cite{chen2020lottery,desai2019evaluating,morcos2019one,mehta2019sparse} investigate the transferability across different datasets (i.e., \textit{dataset transfer}), while other pioneers study the transferability of pre-trained tickets from supervised and self-supervised vision pre-training~\cite{chen2020lottery2} across diverse downstream tasks like detection and segmentation  (i.e., \textit{task transfer}). These two transfer capabilities form the core target of the pre-training / fine-tuning paradigm. In this paper, we take a leap further to meet more practical requirements by designing the concept of double-win tickets. It examines the transferability across different downstream training regimes, including standard and adversarial transfer, data-rich and data-scarce transfer. To our best knowledge, this \textit{training schemes transfer} has never been explored in the LTH literature, offering a new view to analyze beneficial properties of pre-trained tickets.
\paragraph{Adversarial training and robust pre-training.} Deep neural networks are vulnerable to imperceivable adversarial examples~\cite{szegedy2013intriguing}, which limits their applications in security-crucial scenarios. To tackle this limitation, massive defense methods were proposed~\cite{goodfellow2014explaining,kurakin2016adversarial,madry2017towards}, while many of them, except adversarial training \cite{madry2017towards}, were later found to provide false security from obfuscated gradients caused by input transformation~\cite{xu2017feature, liao2018defense, guo2017countering, dziugaite2016study} and randomization~\cite{liu2018adv, liu2018towards, dhillon2018stochastic}. Besides, several works that focus on certified defenses~\cite{cohen2019certified,raghunathan2018semidefinite}, aim to provide a theoretical guarantee of robustness yet lack scalability. 
Nowadays, adversarial training (AT)~\cite{madry2017towards} remains one of the most effective approaches and numerous following works endeavor to improve its performance~\cite{zhang2019theoretically,chen2021robust} and computation efficiency~\cite{shafahi2019adversarial,zhang2020attacks}, while it may suffer from overfitting issues. Particularly, \cite{zhang2019you,free2019, wong2020fast} point out the overfitting phenomenon in several fast adversarial training methods, where sometimes the robust accuracy against a PGD adversary suddenly drops to nearly zero after some training. \cite{andriushchenko2020understanding} suggests it can be mitigated by performing local linearization to the loss landscape in those ``fast" AT. Another reported \textit{robust overfitting}~\cite{rice2020overfitting} seems to raise a completely new challenge for the classical AT (not fast), which can be alleviated by early stopping and smoothening~\cite{chen2021robust}. Meantime, several pioneering efforts have been made to obtain models that are both compact and robust to adversarial attacks~\cite{gui2019model,sehwag2020hydra,fu2021drawing}, spurious features~\cite{zhang2021can}, and input corruptions~\cite{diffenderfer2021winning}

Although the standard pre-training is commonly used in both areas of computer vision~\cite{he2019rethinking,girshick2014rich} and natural language process~\cite{devlin2018bert}, such as the supervised ImageNet and self-supervised BERT~\cite{devlin2018bert} pre-training, there exist only few investigations of robust pre-training.  The work
\cite{Chen_2020_CVPR} for the first time demonstrates that adversarial pre-training can speed up and improve downstream adversarial fine-tuning. Latter works~\cite{jiang2020robust,salman2020adversarially} show extra benefits of enhanced dataset transferability and data efficiency from adversarial pre-training. All the above studies were only conducted with dense networks.


\section{Preliminary}

\vspace{-1mm}
\paragraph{Networks.} Aligned with previous work of pre-trained tickets~\cite{chen2020lottery2}, we consider the official ResNet-50~\cite{he2016deep} as the unpruned dense model, and formulate the output of the network as $f(x; \theta)$, where $x$ is the input images and $\theta\in\mathbb{R}^{d}$ is the network parameters. In the same way, a subnetwork is a network $f(x; m\odot\theta)$\footnote{For simplicity purpose, we use $f(x;\theta)$ to denote a network or its output in different contexts.} with a binary pruning mask $m\in\{0,1\}^{d}$, where $\odot$ is the element-wise product. In our experiment, we sparsify the major part of the dense network, leaving the task-specific classification head out of the scope of pruning.

\vspace{-1mm}
\paragraph{Adversarial training (AT).} The classical AT~\cite{madry2017towards} remains one of the most effective approaches to tackle the vulnerability for small perturbations and build a robust model, in which the standard empirical risk minimization is replaced by a robust optimization, as depicted in~\eqref{eq:minmax}: 
{\small \begin{equation}\label{eq:minmax}
    \min_{\theta}  \mathbb E_{(x, y) \in \mathcal{D}}
    \max_{\left\|\delta\right\|_p \leq \epsilon} \mathcal{L} \big(f(x + \delta; \theta), y \big)
\end{equation}}%
where the perturbation is constrained in an $\ell_p$ norm ball with the radius equals to $\epsilon$, and input data $x$ with its associated label $y$ are sampled from the training set $\mathcal{D}$. To solve the inner maximization problem, projected gradient descent (PGD)~\cite{madry2018towards} is frequently adopted and believed to be the strongest first-order adversary, which works in an iterative fashion as~\eqref{eq:pgd}:
\begin{equation}\label{eq:pgd}
    \delta^{t+1} = \mathrm{proj}_{\mathcal{P}} \Big ( \delta^t + \alpha \cdot \mathrm{sgn} \big ( \nabla_{ x}\mathcal{L}(f(x+\delta^t; \theta),y) \big ) \Big )
\end{equation}
where $\delta^t$ is the generated perturbation, $t$ denotes the number of iterations, $\alpha$ represents the step size, and $\mathrm{sgn}$ is a function that returns the sign of its input. Besides, \cite{wong2020fast} proposes a fast adversarial training method and claimed that adversarial training with Fast Gradient Sign Method (FGSM)~\cite{goodfellow2014explaining}, which is the single-step variant of PGD, can be as effective as PGD-based adversarial training once combined with random initialization. In the following context, we will refer standard empirical risk minimization process to standard training (\textbf{ST}) and robust optimization to adversarial training (\textbf{AT}) or fast adversarial training (\textbf{FAT}) according to the number of PGD steps. We remark that FAT alone may cause the issue of robust catastrophic overfitting~\cite{andriushchenko2020understanding} when the train-time attack strength grows. Thus, an early-stopping policy \cite{rice2020overfitting}, which was also suggested by~\cite{andriushchenko2020understanding},  is adopted to mitigate such catastrophic overfitting.

\vspace{-3mm}
\paragraph{Pruning algorithms.} For a dense neural network $f(x; \theta)$, we adopt the unstructured iterative magnitude pruning (IMP)~\cite{frankle2018lottery,han2015deep} to identify the subnetworks $f(x; m\odot\theta)$, which is a standard option for mining lottery tickets~\cite{frankle2018the}. More precisely, starting from the pre-trained weights $\theta_p$ as initialization, we follow the circle of prune-rewind-retrain to locate subnetworks, in which we prune $p\%$ of the remaining weight with the smallest magnitude and rewind the weights of the subnetwork to their values from $\theta_p$. We repeat the prune-rewind-retrain process until the desired sparsity. 

In our experiments, we choose a precise $p\%=20\%$~\cite{frankle2018the,chen2020lottery2} and consider three initialization: the standard pre-trained\footnote{\scalebox{0.70}{\url{https://pytorch.org/vision/stable/models.html}}} ResNet-50 $\theta_{\mathrm{STD}}$, the PGD-based adversarial pre-trained\footnote{\scalebox{0.70}{\url{https://github.com/microsoft/robust-models-transfer}}} ResNet-50 $\theta_{\mathrm{AT}}$ and fast adversarial pre-trained\footnote{\scalebox{0.65}{\url{https://github.com/locuslab/fast_adversarial/tree/master/ImageNet}}} ResNet-50 by $\theta_{\mathrm{FAT}}$. All the 
models are pre-trained with the classification task on the \textbf{ImageNet} source dataset~\cite{krizhevsky2012imagenet}. It is worthy to mention that all pruning are applied to the source dataset (or pre-training task) only, since our main focus is investigating the mask transferability cross training schemes of subnetworks obtained from pre-training.


\vspace{-3mm}
\paragraph{Downstream datasets, training and evaluation.} After producing subnetworks from the pre-training task on ImageNet by IMP, we implement both standard and adversarial transfer on three downstream datasets: CIFAR-10~\cite{Krizhevsky09}, CIFAR-100~\cite{Krizhevsky09}, and SVHN~\cite{netzer2011reading}. For adversarial training, we train the network against $\ell_{\infty}$ adversary of $10$-steps Projected Gradient Descent (PGD-$10$) with $\epsilon=\frac{8}{255}$ and $\alpha=\frac{2}{255}$. On CIFAR-10/100, we train the network for $100$ epochs with an initial learning rate of $0.1$ and decay by ten times at $50,75$th epoch. As for SVHN, we start from $0.01$ learning rate and decay by a cosine annealing schedule for $80$ epochs. Moreover, an SGD optimizer is adopted with $5\times 10^{-4}$ weight decay and $0.9$ momentum. And we use a batch size of $128$ for all downstream experiments. To evaluate the downstream performance of subnetworks, we report both Standard Testing Accuracy (\textbf{SA}) and Robust Testing Accuracy (\textbf{RA}), which are computed on the original and adversarial perturbed test images respectively. During the inference, we generate the adversarial test images by PGD-$20$ attack with other hyper-parameters kept the same as in training~\cite{chen2021robust}. More details are in Sec.~\ref{sec:more_details}.

\vspace{-3mm}
\paragraph{Double-Win lottery tickets.} Here we introduce formal definitions of our double-win tickets:

$\rhd$ \textit{Matching subnetworks}~\cite{chen2020lottery, chen2020lottery2, frankle2020linear}. A subnetwork $f(x; m\odot\theta)$ is matching for a training algorithm $\mathcal{A}_t^{\mathcal{T}}$ if its performance of evaluation metric $\mathcal{\epsilon}^{\mathcal{T}}$ is no lower than the pre-trained dense network $f(x; \theta_p)$ that trained with the same algorithm $\mathcal{A}_t^{\mathcal{T}}$, namely: 
{\small \begin{equation}
        \mathcal{\epsilon}^{\mathcal{T}} \Big( \mathcal{A}_t^{\mathcal{T}} \big( f(x; m\odot\theta) \big) \Big) \geq \mathcal{\epsilon}^{\mathcal{T}} \Big( \mathcal{A}_t^{\mathcal{T}} \big( f(x; \theta_p) \big) \Big)
    \end{equation}}%
$\rhd$ \textit{Winning Tickets}~\cite{chen2020lottery2, frankle2020linear}. If a subnetwork $f(x; m\odot\theta)$ is matching with $\theta = \theta_p$ for a training algorithm $\mathcal{A}_t^{\mathcal{T}}$, then it is a winning ticket for $\mathcal{A}_t^{\mathcal{T}}$.

$\rhd$ \textit{Double-Win Lottery Tickets}. When a subnetwork $f(x; m\odot\theta)$ is a winning ticket for standard training under metric \textbf{SA} and for adversarial training under both metrics \textbf{SA} and \textbf{RA}, we name it as a double-win lottery ticket, as demonstrated in Table~\ref{tab:settings}.


\begin{table}[t]
\vspace{-4mm}
\caption{Summary of our setups.}
\label{tab:settings}
\centering
\resizebox{1\linewidth}{!}{
\begin{tabular}{l|c|cc}
\toprule
\multicolumn{4}{l}{Source domain: finding subnetworks via pruning with pre-trained weights $\theta_p$} \\ \midrule
\multicolumn{4}{l}{Target domains: evaluating transferability of $f(x; m\odot\theta_p)$} across training schemes \\ \midrule
Training scheme & \multicolumn{1}{c|}{Standard Training} & \multicolumn{2}{c}{Adversarial Training}\\ \midrule
Evaluation metrics & \multicolumn{1}{c|}{SA} & SA & RA \\ \midrule
\textit{Double-Win Tickets} if and only if & winning \cmark & winning \cmark & winning \cmark\\
\bottomrule
\end{tabular}}
\vspace{-6mm}
\end{table}

\begin{figure*}[t]
    \centering
    \includegraphics[width=1\linewidth]{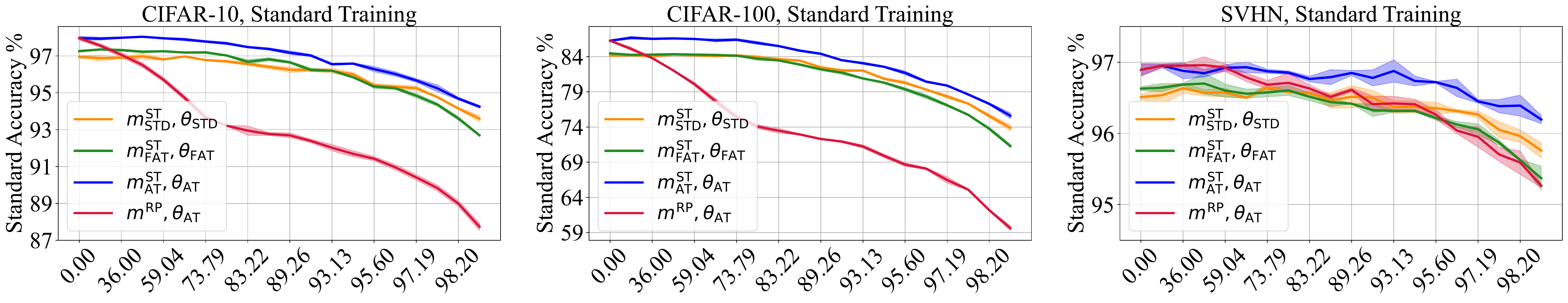}
    \includegraphics[width=1\linewidth]{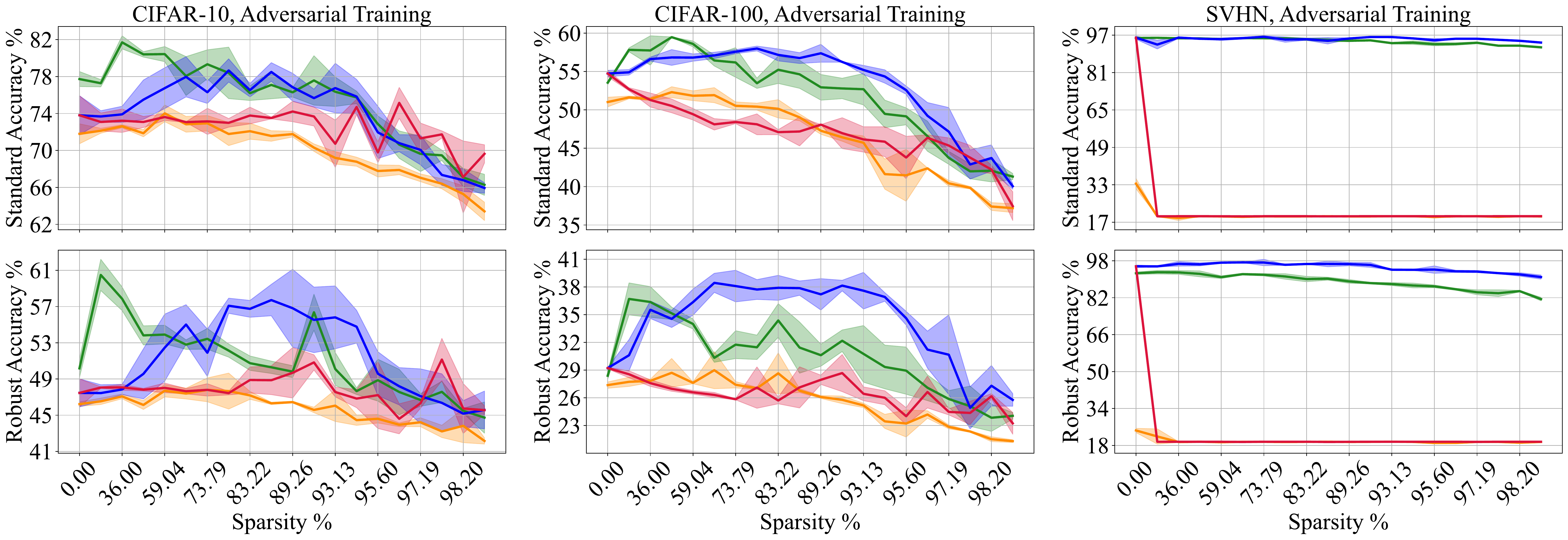}
    \vspace{-8mm}
    \caption{Comparison results of the subnetworks that are fine-tuned on three downstream datasets (i.e., CIFAR-10, CIFAR-100 and SVHN) under both standard and adversarial training regimes. For standard training, we report the standard accuracy; while for adversarial training, both standard and robust accuracy are presented. \textcolor{orange}{Orange}, \textcolor{green}{Green} and \textcolor{blue}{Blue} represent the performance of subnetworks generated from IMP on pre-trained ImageNet classification ($m^{\mathrm{ST}}$) with standard re-training and different pre-trained weights (i.e. standard $\theta_{\mathrm{STD}}$, fast adversarial $\theta_{\mathrm{FAT}}$, and adversarial $\theta_{\mathrm{AT}}$ pre-training, respectively) while \textcolor{red}{Red} stands for random pruning with adversarial pre-trained weight. The solid line and shading area are the mean and standard deviation of standard/robust accuracy.}
    \label{fig:block1}
    \vspace{-2mm}
\end{figure*}

\section{Drawing Double-Win Lottery Tickets from Robust Pre-training}~\label{sec:4}
\vspace{-5mm}

In this section, we evaluate the quality of subnetworks $f(x; m\odot\theta)$ on multiple downstream tasks under both standard and adversarial training regimes. Before that, we extract desired subnetworks via IMP on the ImageNet classification tasks. During the process, the pre-trained weights $\theta_p$ are treated as initialization for rewinding, and standard training (ST) or adversarial training (AT) is adopted for re-training the sparse model on the pre-trained task. In the downstream transferring stage, subnetworks start from mask $m^{\mathcal{P}}_{p}$ and pre-trained weights $\theta_p$, where $p \in \{\mathrm{STD}, \mathrm{FAT}, \mathrm{AT}\}$ stands for \{standard, fast adversarial, adversarial\} pre-training, and the pruning method $\mathcal{P} \in \{\mathrm{ST}, \mathrm{AT}, \mathrm{RP}, \mathrm{OMP}\}$ which represents \{IMP with standard (re-)training, IMP with adversarial (re-)training, random pruning, one-shot magnitude pruning\}~\cite{han2015deep} 
on the pre-training task ($\mathrm{RP}$ or $\mathrm{OMP}$ indicates there is no re-training). In the following content, Section~\ref{sec:4.1} shows the existence of double-win lottery tickets from diverse (robust) pre-training with impressive transfer performance for both standard or adversarial training; Section~\ref{sec:4.2} investigates the effects of standard or adversarial re-training on the quality of derived double-win tickets. All experiments have three independent replicates with different random seeds and the mean results and standard deviation are reported.

\vspace{-1mm}
\subsection{Do Double-Win Lottery Tickets Exist?}\label{sec:4.1}
To begin with, we validate the existence of double-win lottery tickets drawn from diverse (robust) pre-training and source ImageNet dataset. We consider the sparsity masks from IMP with standard re-training $m^{\mathrm{ST}}$ and random pruning $m^{\mathrm{RP}}$ on the pre-training task, 
together with three different pre-trained weights, i.e. standard $\theta_{\mathrm{STD}}$, fast adversarial $\theta_{\mathrm{FAT}}$, and adversarial weights $\theta_{\mathrm{AT}}$. As demonstrated in Figure~\ref{fig:block1}, we adopt two downstream fine-tuning receipts, i.e., standard training (report SA) and adversarial training (report SA/RA)), simultaneously. \textbf{Note that all presented numbers here are subnetwork's sparsity levels.} Several consistent observations can be drawn:

\begin{figure*}[t]
    \centering
    \includegraphics[width=1\linewidth]{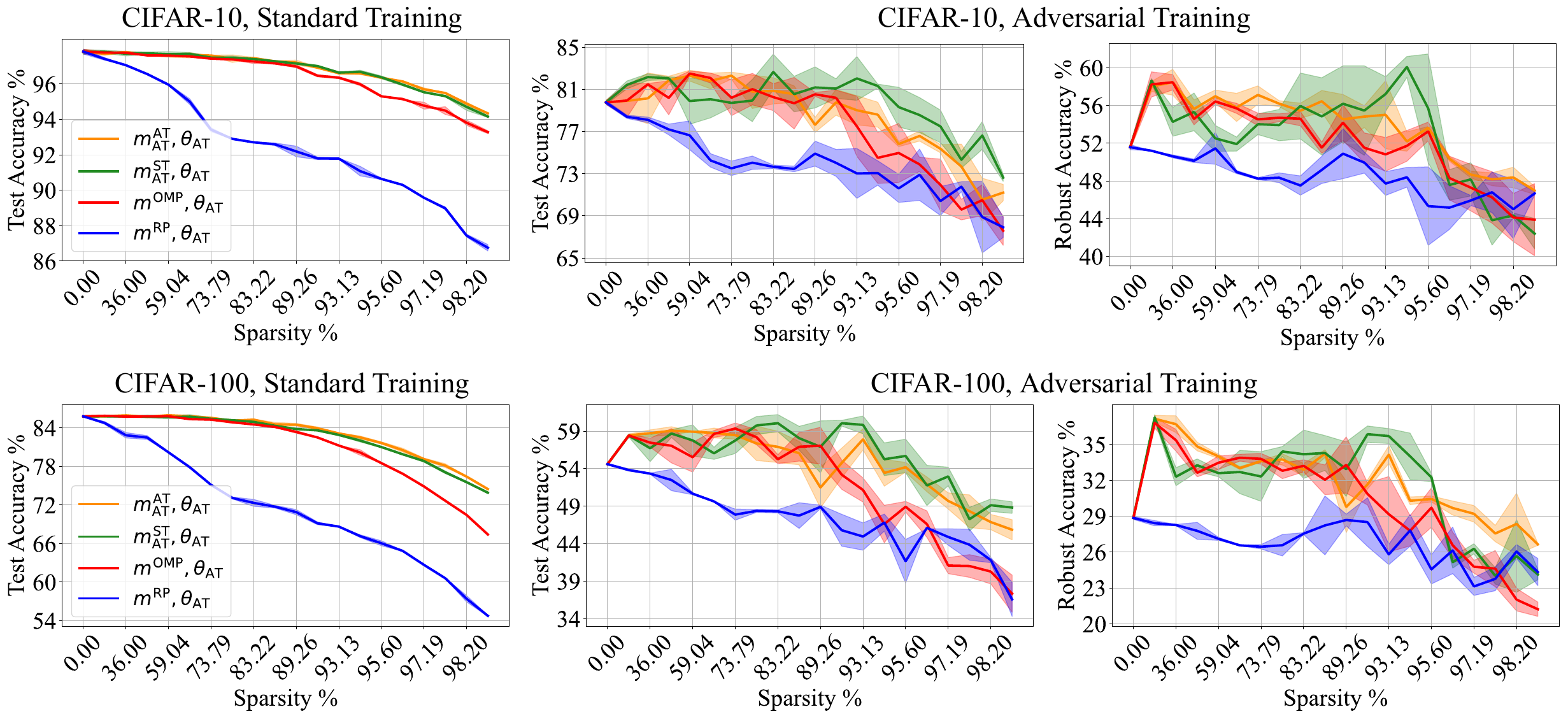}
    \vspace{-8mm}
    \caption{{\small Comparison results of the subnetworks that are independently trained on three downstream datasets (i.e., CIFAR-10, CIFAR-100, SVHN) under both standard and adversarial training regimes. For standard training, we report the standard accuracy; while for adversarial training, both standard and robust accuracy are presented. \textcolor{orange}{Orange}, \textcolor{green}{Green}, \textcolor{red}{red} and \textcolor{blue}{Blue} represent the performance of subnetworks generated by IMP with standard re-training ($m^{\mathrm{ST}}_{\mathrm{AT}}$), adversarial re-training ($m^{\mathrm{AT}}_{\mathrm{AT}}$) on ImageNet classification task, one shot magnitude pruning ($m^{\mathrm{OMP}}$) and random pruning ($m^{\mathrm{RP}}$), together with the adversarial pre-training ($\theta_{\mathrm{AT}}$) as initialization. The solid line and shading area are the mean and standard deviation of standard/robust accuracy.}}
    \label{fig:train_regime}
    \vspace{-1mm}
\end{figure*}

\begin{itemize}
    \vspace{-1mm}
    \item [\ding{172}] Double-win lottery tickets generally exist from various pre-training, showing unimpaired performance on diverse downstream tasks for both standard and adversarial transfer.  
    To account for fluctuations, we consider the performance of subnetworks is \textit{matching} when it's within one standard deviation of the unpruned dense network. 
    The \textbf{extreme sparsity levels} of subnetworks drawn from \{$\theta_{\mathrm{STD}}$, $\theta_{\mathrm{FAT}}$, $\theta_{\mathrm{AT}}$,\} are \{$89.26\%$, $89.26\%$, $91.41\%$\}, \{$73.79\%$, $79.03\%$, $83.22\%$\}, \{$0.00\%$, $79.03\%$, $96.48\%$\} with matching or even superior standard and robust performance under both training regimes (standard and adversarial) on CIFAR-10, CIFAR-100 and SVHN, respectively. All these double-win tickets surpass randomly pruned subnetwork by a significant performance margin. It demonstrates the superior performance of double-win tickets is not only from reduced parameter counts but also credits to the located sparse structural patterns.
    \vspace{-1mm}
    \item [\ding{173}] Subnetworks identified from adversarial pre-training consistently outperform the ones from fast adversarial and standard pre-training across all three downstream classification tasks, which is aligned with the result in~\cite{salman2020adversarially}. Taking the extreme sparsity as an indicator, the adversarial pre-training finds double-win lottery tickets to the extreme sparsity of $83.22\% \sim 96.48\%$ while fast adversarial, standard pre-training reach the extreme sparsity level of $20.00\% \sim 89.26\%$ and $0.00\% \sim 89.26\%$. This suggests that the adversarial pre-trained model can serve as a desirable starting point for locating high-quality double-win tickets to cover both standard and adversarial downstream transferability. Note that here all downstream transferring can access full training data, i.e., \textit{data-rich} fine-tuning.
    \vspace{-1mm}
    \item [\ding{174}] Along with the increase of sparsity, we notice that the performance improvements from adversarial pre-training $\theta_{\mathrm{AT}}$ ($i$) remain stable in the standard transfer (the first row in Figure~\ref{fig:block1}) even at extreme sparsity like $98.56\%$; ($ii$) first increase then diminish in adversarial transfer after $95.60\%$ sparsity. It suggests that double-win tickets from adversarial pre-training are more sensitive to the aggressive sparsity in the scenario of adversarial transfer learning. 
    \vspace{-1mm}
    \item [\ding{175}] The comparison results among different pre-training varies with the training regime of downstream tasks. Take the result on CIFAR-100 as an example, the subnetworks drawn from fast adversarial pre-training shows superior performance than the ones from standard training in the range of sparsity level from $0.00\% \sim 98.56\%$ under adversarial training. While for the standard training regime, fast adversarial and standard pre-training locate subnetworks with similar performance across the sparsity level from $0.00\%$ to $95.60\%$. The inferior performance of standard pre-training suggests that the vanilla lottery tickets that only focus on the standard training regime and use standard test accuracy as the only evaluation metric, is insufficient in practical security-crucial scenarios. Thus we take adversarial transfer into consideration and propose the concept of double-win lottery tickets to improve the original LTH.
\end{itemize}





\begin{figure*}[t]
    \centering
    \includegraphics[width=0.92\linewidth]{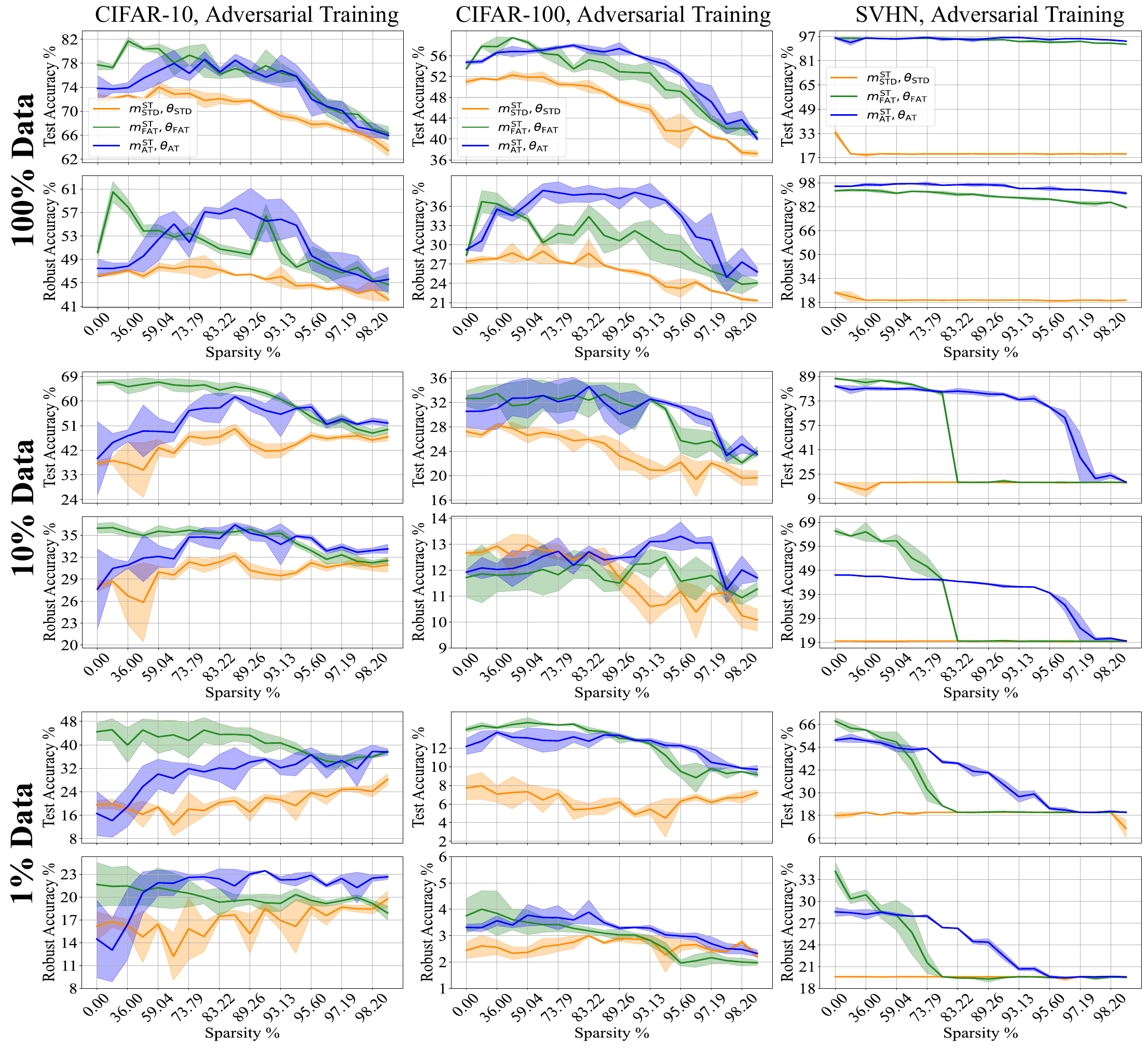}
    \vspace{-4mm}
    \caption{{\small Data-efficient transfer results of double-win tickets from adversarial pre-training on three downstream datasets (i.e. CIFAR-10, CIFAR-100, and SVHN) with $100$\%, $10$\% and $1$\% training data. Both standard and robust accuracy are reported. \textcolor{orange}{Orange}, \textcolor{green}{Green} and \textcolor{blue}{Blue} represent the performance of subnetworks located from IMP together with standard training on ImageNet classification ($m^{\mathrm{ST}}$) with different pre-trained weights (i.e. standard $\theta_{\mathrm{STD}}$, fast adversarial $\theta_{\mathrm{FAT}}$, and adversarial pre-training $\theta_{\mathrm{AT}}$, respectively). The solid line and shading area are the mean and standard deviation of standard/robust accuracy.}}
    \label{fig:data_efficience}
    \vspace{-5mm}
\end{figure*}

\vspace{-1mm}
\subsection{Do training regimes on source domain affect the located subnetworks?}\label{sec:4.2}

During the ticket finding on pre-trained tasks, we can adopt standard re-training and adversarial re-training after each IMP pruning process. Intuitively, adversarial re-training should be able to maintain more information from adversarial pre-training, and lead to better transfer performance on downstream tasks~\cite{salman2020adversarially}. However, our experiment results surprisingly challenge this ``common sense". Specifically, we choose the adversarial pre-trained weight ($\theta_{\mathrm{AT}}$\footnote{The official robust model ($\ell_{\infty}$ adversary with $\epsilon=\frac{8}{255}$ and $\alpha=\frac{2}{255}$) on ImageNet zoo at \scalebox{0.61}{\url{https://github.com/MadryLab/robustness}}.}) as the initialization and compare four types of pruning and re-training methods on the pre-trained tasks, i.e., IMP with standard training ($m^{\mathrm{ST}}_{\mathrm{AT}}$), IMP with adversarial training ($m^{\mathrm{AT}}_{\mathrm{AT}}$), one-shot magnitude pruning (OMP) ($m^{\mathrm{OMP}}$) and random pruning ($m^{\mathrm{RP}}$). 

As shown in Figure~\ref{fig:train_regime}, the extreme sparsity of double-win lottery tickets on \{CIFAR-10, CIFAR-100\} is ($91.41$\%, $83.22$\%), ($91.41$\%, $83.22$\%), ($89.26$\%, $79.03$\%), ($20$\%, $0$\%) for ($m^{\mathrm{AT}}_{\mathrm{AT}}, \theta_{\mathrm{AT}}$), ($m^{\mathrm{ST}}_{\mathrm{AT}}, \theta_{\mathrm{AT}}$), ($m^{\mathrm{OMP}}, \theta_{\mathrm{AT}}$) and ($m^{\mathrm{RP}}, \theta_{\mathrm{AT}}$), respectively. IMP with standard and adversarial training shows similar performance, and both of them are better than OMP and random pruning. It suggests that the re-training regimes during IMP doesn't make an significant impact on the downstream transferablity of subnetworks. Due to the heavy computational cost of adversarial training and the inferior performance of OMP, we consider IMP with standard training as our major pruning method and investigate the data efficiency property of subnetworks in following sections.



\section{Double-Win Tickets with Robust Pre-training Enables Data-Efficient Transfer}\label{sec:5}


In this section, we further exploit the practical benefits of double-win lottery tickets by assessing the data-efficient transferability under limited training data schemes (e.g., $1$\% and $10$\%). All subnetworks are drawn from IMP with standard re-training on the pre-training task. We consider three different pre-trained weights (i.e., standard $\theta_{\mathrm{STD}}$, fast adversarial $\theta_{\mathrm{FAT}}$, and adversarial pre-training $\theta_{\mathrm{AT}}$). The results are included in Figure~\ref{fig:data_efficience}, from which we find:
\begin{itemize}
    \vspace{-2mm}
    \item [\ding{172}] $\theta_{\mathrm{FAT}}$ and $\theta_{\mathrm{AT}}$ significantly outperform $\theta_{\mathrm{STD}}$ on both data-rich and data-scarce transfer for all three datasets. It evidences that the robust pre-training improves data-efficient transfer. While on the challenging SVHN downstream dataset with limited training data (i.e., $10\%$ and $1\%$), the performance of $\theta_{\mathrm{FAT}}$ and $\theta_{\mathrm{AT}}$ degrades to $\theta_{\mathrm{STD}}$'s level at large sparsity $83.22\%$ and $97.19\%$ respectively.
    \vspace{-1mm}
    \item [\ding{173}] For data-limited transferring, sparse tickets derived from robust pre-training \{$\theta_{\mathrm{FAT}}$,$\theta_{\mathrm{AT}}$\} surpass their dense counterpart by up to \{$0.75\%$,$22.53\%$\} SA and \{$0.79\%$,$8.97\%$\} RA, which indicates the enhanced data-efficiency also comes from appropriate sparse structures. The consistent robustness gains under unseen transfer attacks in Section~\ref{sec:more_res}, also exclude the possibility of obfuscated gradients.
    \vspace{-1mm}
    \item [\ding{174}] In general, when subnetworks are trained with only $10\%$ or $1\%$ training samples available, those drawn from fast adversarial pre-trained weight $\theta_{\mathrm{FAT}}$ show superior performance at middle sparsity levels, with performance improvements up to \{$30.97\%$, $2.42\%$, $10.05\%$\} SA and \{$8.36\%$, $0.70\%$, $18.49\%$\} RA compared with $\theta_{\mathrm{AT}}$ for CIFAR-10, CIFAR-100 and SVHN, respectively. But with the increase of sparsity, adversarial pre-trained weight $\theta_{\mathrm{AT}}$ overtakes $\theta_{\mathrm{FAT}}$ and dominates the larger sparsity range.
\end{itemize}

\begin{figure}[t]
    \centering
    \vspace{-1mm}
    \includegraphics[width=0.85\linewidth]{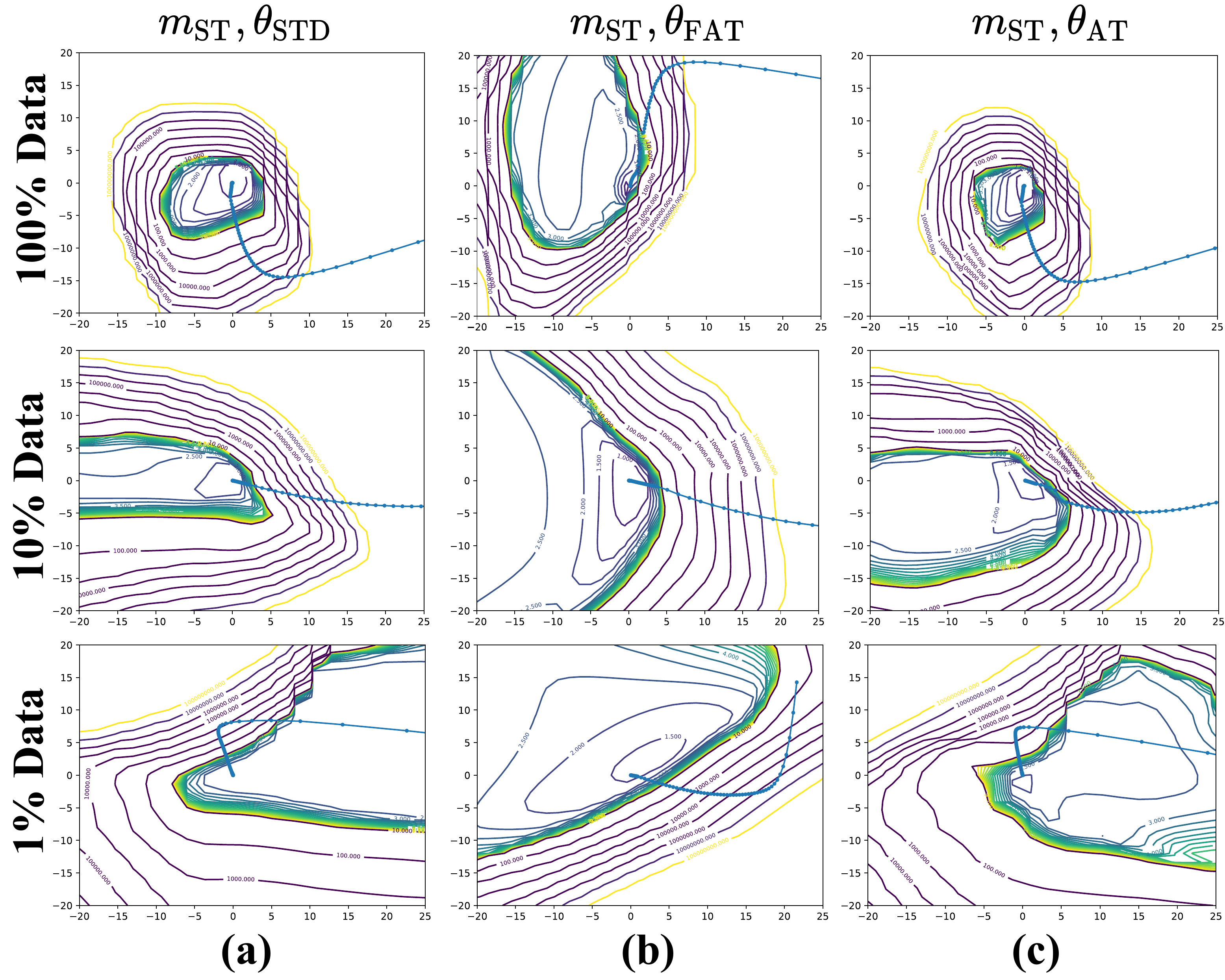}
    \vspace{-4mm}
    \caption{Visualization of loss contours and training trajectories of subnetworks located by IMP with standard re-training $m^{\mathrm{ST}}$ at $73.79\%$ sparsity. Each subnetwork is adversarial trained with $100\%$, $10\%$ or $1\%$ training data on CIFAR-10. We compare three pre-training (i.e., standard $\theta_{\mathrm{STD}}$, fast adversarial $\theta_{\mathrm{FAT}}$, and adversarial pre-training $\theta_{\mathrm{AT}}$). The original test set is used.}
    \label{fig:loss_land}
    \vspace{-5mm}
\end{figure}

\vspace{-2mm}
To understand the counter-intuitive results that subnetworks from weak robust pre-training $\theta_{\mathrm{FAT}}$ perform better than the ones from strong robust pre-training $\theta_{\mathrm{AT}}$ at middle sparsity levels such as $73.79\%$ particularly for data-limited transferring, we visualize the training trajectories along with loss landscapes through tools in~\cite{visualloss}. We take robustified subnetworks with $73.79\%$ sparsity on CIFAR-10 as an example. As shown in Figure~\ref{fig:loss_land} and~\ref{fig:loss_land_adv}, for the results on the original test data (columns: a,b,c), the loss contour of $\theta_{\mathrm{FAT}}$ is smoother/flatter than $\theta_{\mathrm{AT}}$ and $\theta_{\mathrm{STD}}$, i.e., the basin with converged minimum has larger area in terms of the same level of loss like the $2.000$ contour in the middle row's plots of Figure~\ref{fig:loss_land}. A smoother/flatter loss surface is often believed to indicate enhanced standard~\cite{keskar2017large, he2019asymmetric} and robust generalization~\cite{wu2020revisiting,hein2017formal}. It offers a possible explanation of $\theta_{\mathrm{FAT}}$'s superior performance to $\theta_{\mathrm{AT}}$'s by up to $
8.98\%$ and $9.62\%$ SA improvements for data-limited transferring with $10\%$ and $1\%$ training samples. Moreover, loss geometric on attacked test data (Fig.~\ref{fig:loss_land_adv}) reveals similar conclusions. 

 \vspace{-2mm}
\section{Analyzing Properties of Double-Win Tickets}
 \vspace{-1.5mm}
\paragraph{Relative similarity.} In Fig.~\ref{fig:relative_similar}, we report the relative similarity (i.e., $\frac{|m_i \cap m_j|}{|m_i \cup m_j|}$) between binary pruning masks $m_i$ and $m_j$, which denotes the degree of overlapping in sparse patterns located from different pre-trained models. We observe that subnetworks from different pre-training has distinct sparse patterns. Specifically, the relative similarity is less than $20.00$\% when the sparsity of subnetworks reaches $73.79$\% and the more sparsified, the larger differences arise.

\begin{figure}[t]
    \centering
    \includegraphics[width=0.92\linewidth]{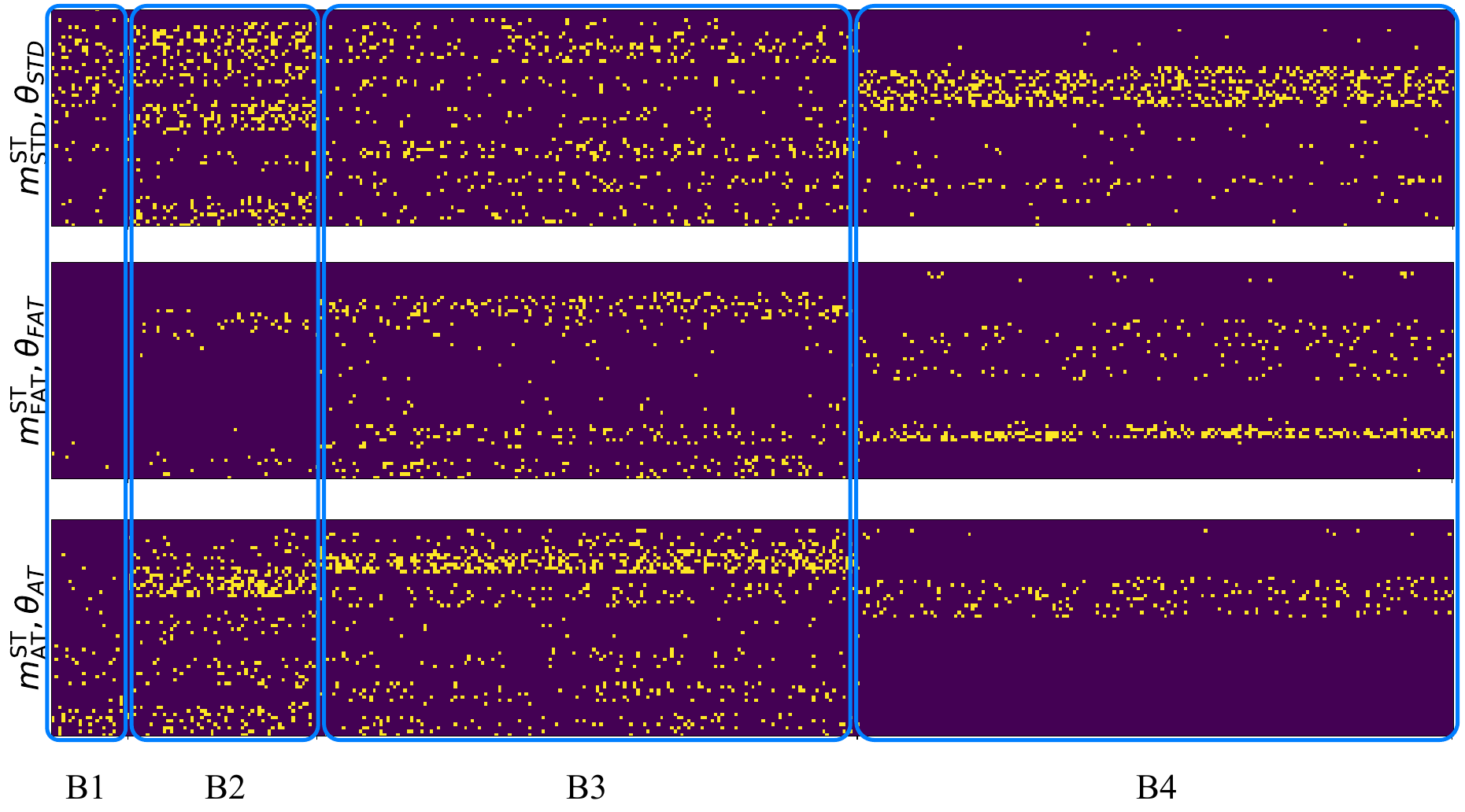}
    \vspace{-4mm}
    \caption{{\small Kernel-wise heatmap visualizations of sparse masks drawn from three different pre-training, i.e., $m^{\mathrm{ST}}_{\mathrm{STD}}$, $m^{\mathrm{ST}}_{\mathrm{FAT}}$, and $m^{\mathrm{ST}}_{\mathrm{AT}}$ at $97.19\%$ sparsity. The bright dots (\textcolor{yellow}{$\bullet$}) are the completely pruned (zero) kernels and the dark dots (\textcolor[RGB]{202,12,22}{$\bullet$}) stand for the kernels with at least one remaining weight. $\mathrm{B}1 \sim \mathrm{B}4$ represent the four residual blocks in ResNet-50.}}
    \label{fig:heatmap}
    \vspace{-3mm}
\end{figure}

\vspace{-3mm}
\paragraph{Structural patterns.} Meanwhile, we calculate the number of completely pruned (zero) kernels and visualize the kernel-wise heatmap of subnetworks with an extreme sparsity of $97.19$\%. As depicted in Figure~\ref{fig:heatmap}, the subnetworks from the standard pre-trained model have the largest number of zero kernels, which roughly reveals the most clustered sparse patterns. And the subnetworks from robust pre-training are less clustered, especially for $\theta_{\mathrm{FAT}}$. We notice that these zero kernels are mainly distributed in the front/later residual blocks for subnetworks from $\theta_{\mathrm{AT}}$/$\theta_{\mathrm{FAT}}$, where they scatter evenly across all blocks. Typically, subnetworks with more zero kernels may have a stronger potential for hardware speedup~\cite{elsen2020fast}.

 \vspace{-1mm}
\section{Conclusion and Limitation}
 \vspace{-1mm}
In this paper, we examine the lottery tickets hypothesis in a more rigorous and practical scenario, which asks for competitive transferability across both standard and adversarial downstream training regimes. We name these intriguing subnetworks as double-win lottery tickets. Extensive results reveal that double-win matching subnetworks derived from robust pre-training enjoy superior performance and enhanced data efficiency during transfer learning. However, the current investigations are only demonstrated in computer vision, and we leave the exploration in other fields such as natural language processing to future work.

\vspace{-1mm}
{\small \section*{Acknowledgment}
\vspace{-0.5em}
Z.W. is in part supported by an NSF RTML project (\#2053279).}


\bibliography{DWLTH}
\bibliographystyle{icml2022}

\newpage
\appendix
\renewcommand{\thepage}{A\arabic{page}}  
\renewcommand{\thesection}{A\arabic{section}}   
\renewcommand{\thetable}{A\arabic{table}}   
\renewcommand{\thefigure}{A\arabic{figure}}

\clearpage

\section{More Implementation Details} \label{sec:more_details}
To identify subnetworks in the pre-trained models, we consider both standard and adversarial re-training for IMP, in which we remove $20\%$ parameters with the lowest magnitude for each pruning step and fine-tune the network for $30$ epochs with a fixed learning rate of $5 \times 10^{-4}$. And we use an SGD optimizer with the weight decay and momentum kept to $1 \times 10^{-4}$ and $0.9$, respectively. The batch size equals $2048$ for all experiments of IMP on the pre-training task with ImageNet. And for adversarial training, we apply PGD-$3$ with $\epsilon=\frac{8}{255}$ and $\alpha=\frac{2}{255}$ against the $\ell_{\infty}$ adversary.

\vspace{-2mm}
\section{More Experiments Results} \label{sec:more_res}
\vspace{-4mm}

\vspace{-2mm}
\paragraph{Excluding obfuscated gradients.}
\begin{table}[!htb]
  \centering
  \caption{Transfer attack performance of subnetworks located from adversarial pretraining $\theta_{\mathrm{AT}}$ through IMP with standard re-training $m^{\mathrm{ST}}$. And the subnetworks are trained with PGD-$10$ on CIFAR-10. We report the accuracy on attacked test sets, which are generated from an unseen robust model, together with the vanilla robust accuracy.}
  \resizebox{0.95\linewidth}{!}{
    \begin{tabular}{c|c|c}
    \toprule
    Sparsity (\%) & Transfer Attack Accuracy (\%) & Robust Accuracy (\%) \\ \midrule
    $0$ & $55.54$ & $47.11$ \\
    $89.26$ & $62.79$ & $56.00$ \\
    $93.13$ & $60.09$ & $60.05$ \\
    $95.60$ & $56.49$ & $50.85$ \\
    \bottomrule
    \end{tabular}}
    \vspace{-1mm}
  \label{tab:transfer_attack}%
\end{table}
Table~\ref{tab:transfer_attack} demonstrates that the sparse subnetworks consistently outperform the dense counterpart under transfer attack from an unseen robust model, which is aligned with the vanilla robust accuracy. This piece of evidence excludes the possibility of gradient masking for our obtain RA improvements.

\begin{figure}[!htb]
    \centering
    \vspace{-4mm}
    \includegraphics[width=1\linewidth]{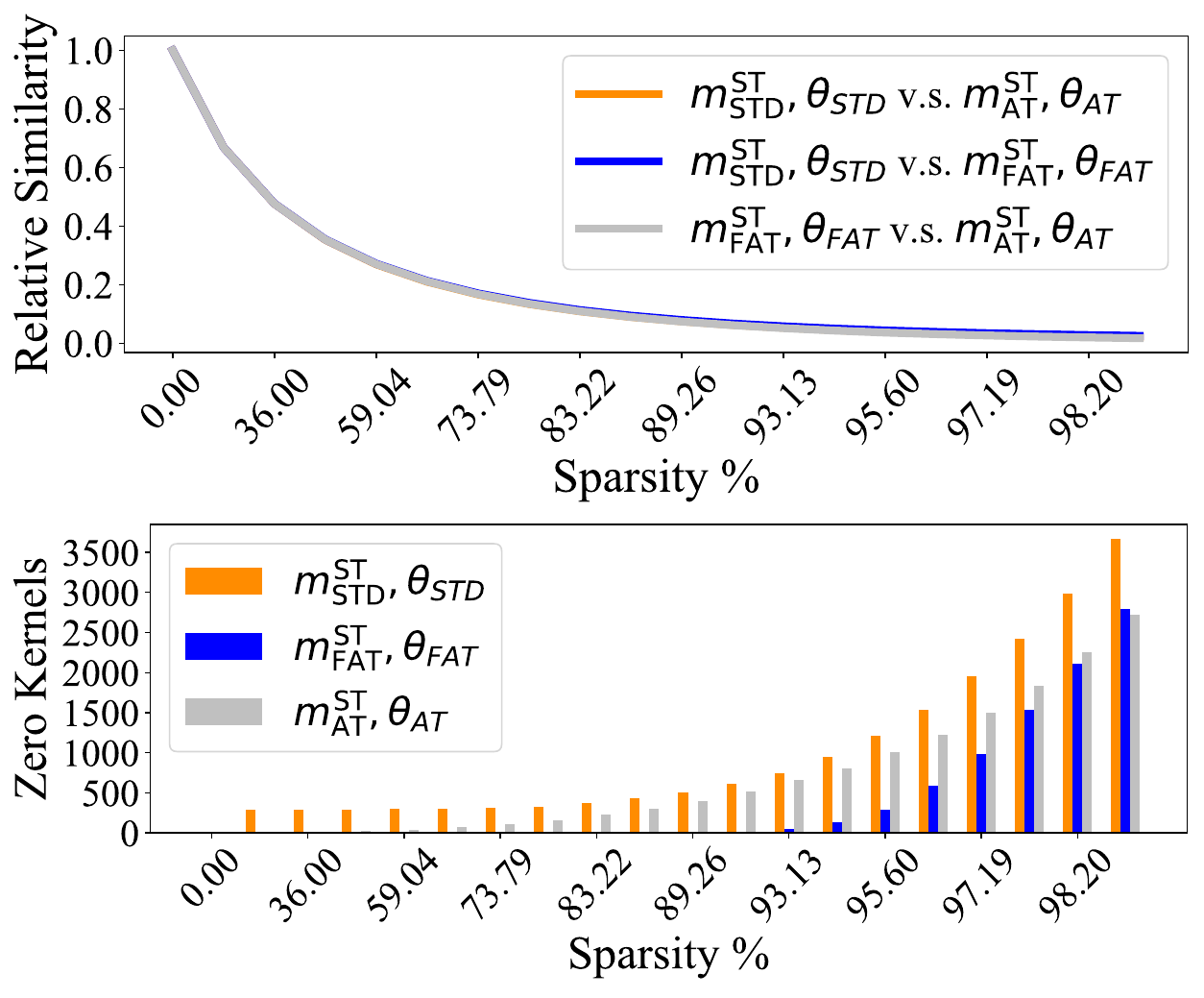}
    \vspace{-10mm}
    \caption{\small The  statistic  of  subnetworks  drawn  from  three  different  pre-training, i.e. ($m^{\mathrm{ST}}_{\mathrm{STD}}, \theta_{\mathrm{STD}}$), ($m^{\mathrm{ST}}_{\mathrm{FAT}}, \theta_{\mathrm{FAT}}$) and ($m^{\mathrm{ST}}_{\mathrm{AT}}, \theta_{\mathrm{AT}}$). (Top): The relative mask similarity between subnetworks from different pre-training. (Bottom): The number of completely pruned (zero) kernels in these subnetworks.}
    \label{fig:relative_similar}
    \vspace{-4mm}
\end{figure}

\paragraph{Relative similarity.}
To measure the overlapping level in sparse patterns drawn from different pre-trained models, we adopt the relative similarity (i.e., $\frac{|m_i \cap m_j|}{|m_i \cup m_j|}$) between binary pruning masks $m_i$ and $m_j$. As shown in Fig.~\ref{fig:relative_similar}, subnetworks from different pre-training share remarkably heterogeneous sparse structures. For instance, the relative similarity is less than $20.00$\% when the sparsity of subnetworks reaches $73.79$\% and the more sparsified, the larger differences arise.

\begin{figure}[t]
    \centering
    \includegraphics[width=0.9\linewidth]{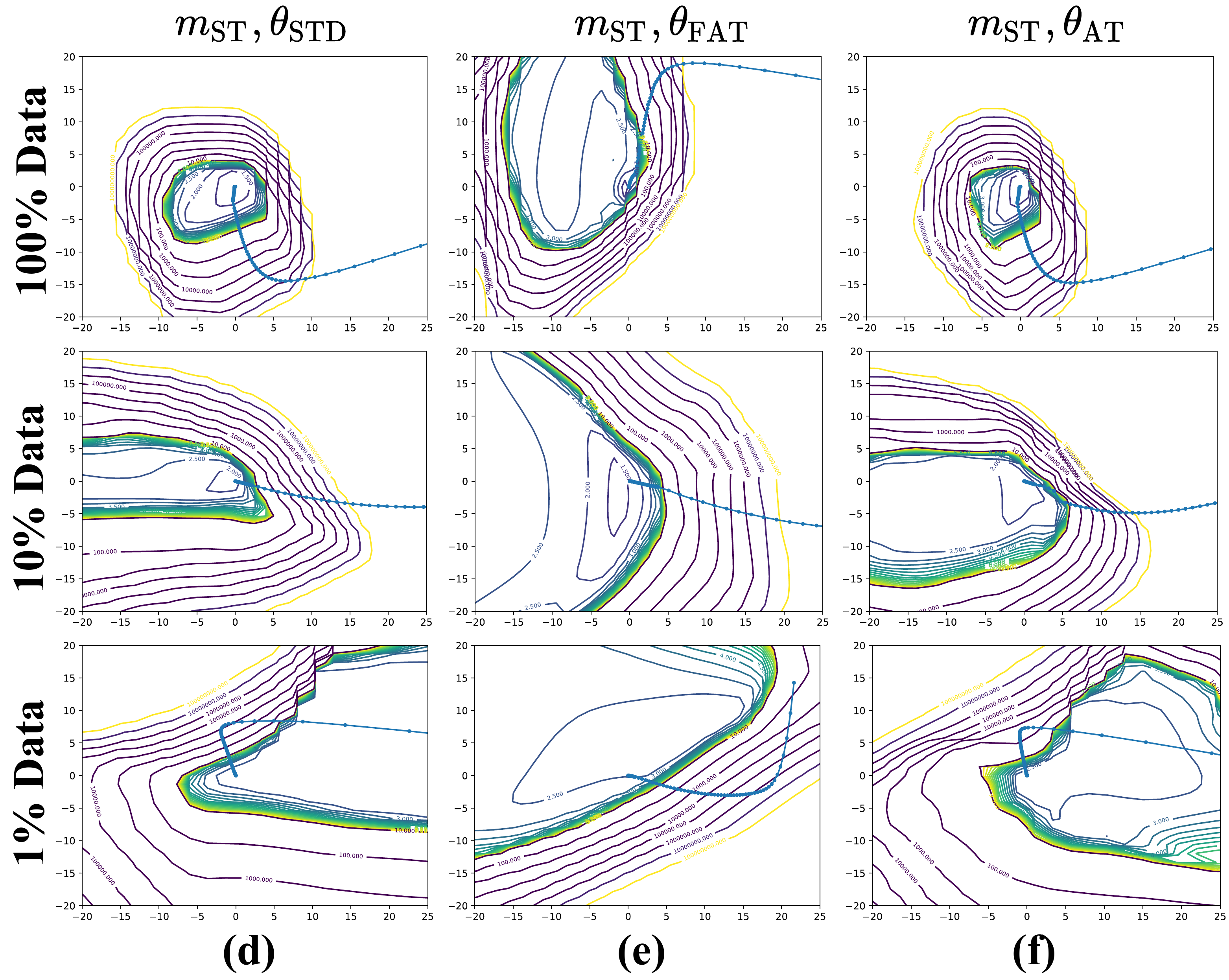}
    \vspace{-4mm}
    \caption{Visualization of loss contours and training trajectories of subnetworks located by IMP with standard re-training $m^{\mathrm{ST}}$ at $73.79\%$ sparsity. Each subnetwork is adversarial trained with $100\%$, $10\%$ and $1\%$ training data on CIFAR-10, respectively. We compare three pre-training (i.e., standard $\theta_{\mathrm{STD}}$, fast adversarial $\theta_{\mathrm{FAT}}$, and adversarial pre-training $\theta_{\mathrm{AT}}$). Columns (d,e,f) stand for the results on attacked test data by PGD-$20$.}
    \label{fig:loss_land_adv}
    \vspace{-3mm}
\end{figure}

\vspace{-2mm}
\paragraph{Extra loss surface visualizations.} As shown in Figure~\ref{fig:loss_land_adv}, consistent observations with Figure~\ref{fig:loss_land} can be drawn.

\begin{figure}[!ht] 
\centering
\vspace{-2mm}
\includegraphics[width=1\linewidth]{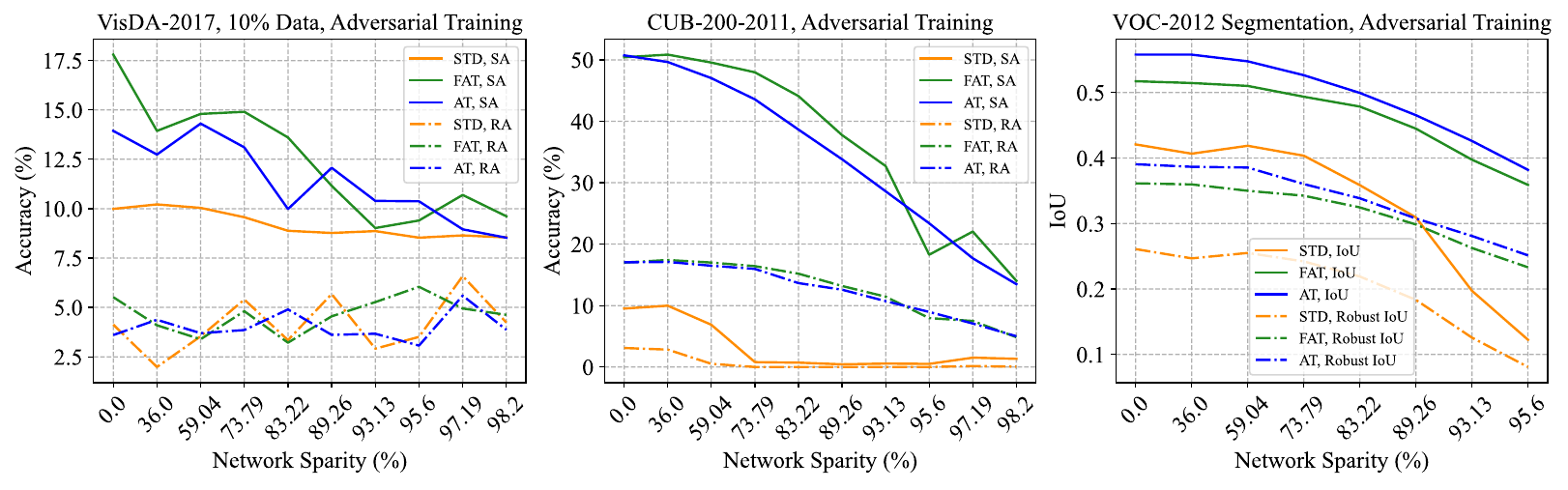}
\vspace{-8mm}
\caption{Results on more datasets and tasks.}
\label{fig:more_res}
\vspace{-4mm}
\end{figure}

\vspace{-2mm}
\paragraph{More datasets and tasks.} We conduct additional experiments of classification on ($i$) CUB-200 birds (more classes and higher resolution), ($ii$) VisDA17 ($4\sim5$ times bigger than CIFAR), and ($iii$) instance segmentation on VOC. Results of $60$ sparse models are collected in Figure~\ref{fig:more_res}, showing consistent conclusion that robust pre-training helps. 

\section{Boarder Impact}
Although our work makes great contributions to efficient machine learning and security-critical applications, it still has potential negative social impacts when it is abused by malicious attackers. Specifically, our methods may speed up and robustify attackers' harmful algorithms or software. One possible solution is to issue a license and limit the blind distribution of our proposals.
\end{document}